\documentclass[11pt]{article}
\usepackage[utf8]{inputenc}
\usepackage[margin=1in]{geometry}

\usepackage[colorlinks=true,citecolor=blue]{hyperref}

\usepackage{microtype}
\usepackage{graphicx}
\usepackage{booktabs} 
\usepackage{natbib}
\setcitestyle{open={(},close={)}}

\usepackage{times}
\usepackage{latexsym}
\usepackage[T1]{fontenc}
\usepackage[utf8]{inputenc}

\usepackage{microtype}

\usepackage{adjustbox}
\usepackage{booktabs}
\usepackage{tabularx}
\usepackage{multirow}
\usepackage{amsmath}
\usepackage{xcolor}
\usepackage{algorithm}
\usepackage{algorithmic}
\usepackage{enumitem}
\usepackage{caption}
\usepackage{subcaption}
\usepackage{arydshln}
\usepackage{amsmath}
\usepackage{amssymb}
\usepackage{dsfont}

\begin{document}

\title{\huge Prompting through Prototype: A Prototype-based Prompt Learning on Pretrained Vision-Language Models}

\author{\vspace{0.3in}\\\textbf{Yue Zhang, Hongliang Fei, Dingcheng Li, Tan Yu, Ping Li} \\\\
Cognitive Computing Lab\\
Baidu Research\\
10900 NE 8th St. Bellevue, WA 98004, USA\\\\
  \texttt{\small\{yuezhang030, feihongliang0, dingchengl, tanyu1503, pingli98\}@gmail.com}
}

\date{\vspace{0.5in}}
\maketitle

\begin{abstract}\vspace{0.3in}
\noindent Prompt learning is a new learning paradigm which reformulates downstream tasks as similar pretraining tasks on pretrained models by leveraging textual prompts. Recent works have demonstrated that prompt learning is particularly useful for few-shot learning, where there is limited training data. Depending on the granularity of prompts, those methods can be roughly divided into task-level prompting and instance-level prompting. Task-level prompting methods learn one universal prompt for all input samples, which is efficient but ineffective to capture subtle differences among different classes. Instance-level prompting methods learn a specific prompt for each input, though effective but inefficient.  
In this work, we develop a novel prototype-based prompt learning method to overcome the above limitations. In particular, we focus on few-shot image recognition tasks on pretrained vision-language models (PVLMs) and develop a method of prompting through prototype (\texttt{PTP}), where we define $K$ image prototypes and $K$ prompt prototypes. In \texttt{PTP}, the image prototype represents a centroid of a certain image cluster in the latent space and a prompt prototype is defined as a soft prompt in the continuous space. The similarity between a query image and an image prototype determines how much this prediction relies on the corresponding prompt prototype. Hence, in \texttt{PTP}, similar images will utilize similar prompting ways. Through extensive experiments on seven real-world benchmarks, we show that \texttt{PTP} is an effective method to leverage the latent knowledge and adaptive to various PVLMs. Moreover, through detailed analysis, we discuss pros and cons for prompt learning and parameter-efficient fine-tuning under the context of few-shot learning. 
\end{abstract}

\newpage

\section{Introduction}
Prompt learning~\citep{li2021prefix, gao2021making, sanh2022multitask} is a new  paradigm to reformulate downstream tasks as similar pretraining tasks on pretrained language models (PLMs) with the help of a textual prompt. Compared with the conventional ``pre-train, fine-tuning'' paradigm, prompt learning is particularly useful~for few-shot learning, where there is no sufficient training data to fine-tune the whole pre-trained~model. Recently, light-weight but effective prompt learning methods have been~developed in various few-shot learning tasks~\citep{schick2021few, gao2021making, shin2020autoprompt} in natural language processing (NLP), such as few-shot sentiment analysis and natural~language~inference. 

With the success of prompt learning in NLP, it is natural to generalize prompt learning to pretrained vision-language models (PVLMs)~~\citep{radford2021learning, kim2021vilt, jin2022good, zhou2022learning, tsimpoukelli2021multimodal, liang2022modular, sanh2022multitask} for vision-language tasks. In this work, we especially focus on exploring few-shot image recognition tasks in the prompt learning paradigm, which has not been fully explored in the prompt learning research area.
The motivation originates from the fact that PVLMs, such as CLIP~\citep{radford2021learning} and ViLT~\citep{kim2021vilt}, are pre-trained with image-text matching and masked language modeling (MLM) style tasks on images and their aligned descriptions. For the image recognition task, where class labels have a textual form (e.g. ``faces'', ``Hummer SUV''),  they can be converted into image-text matching~tasks. For example, one simple manual-craft prompt template could be ``a photo of a [CLASS]'', where [CLASS] will be replaced by any candidate category name. The PVLM matches~the query image with all the prompted candidate category names, and chooses the one with the highest~matching~score. 

Similar to NLP, the essence of prompt learning for PVLM is designing the most appropriate prompts for the downstream tasks. The latest methods to construct prompts include, i) manual-craft prompts~\citep{petroni2019language, jin2022good}, where researchers manually create intuitive templates based on human introspection; ii) automatically searched prompts~\citep{shin2020autoprompt, zhong2021factual, zhou2022learning}, where researchers search over discrete input token space or continuous embedding space for prompts that elicit correct predictions in the training set; iii) instance-level prompt learning~\citep{zhou2022cocoop, rao2022denseclip, jin2022instance}, where instead of learning one universal prompt that works for all the input, they learn instance-level prompts conditional on the given input. Although manually written prompts are interpretable, they are limited by the manual effort, and might not be optimal for eliciting correct predictions. The automated approaches overcome the limitations of manual prompts by training a statistical model, but they learn one universal prompt for each task, which may result in sub-optimal prompts. Instance-level prompt learning methods learn different prompts conditional on the given inputs, however, they usually need to maintain a complex neural module mapping the inputs into prompts, which makes them work poorly on few-shot learning settings. 

Meanwhile, besides prompt learning on PVLMs, researchers are also exploring parameter-efficient fine-tuning methods for few-shot learning, such as linear probing~\citep{tian2020rethinking}, Adaptor~\citep{houlsby2019parameter}, Bitfit~\citep{zaken2022bitfit} and Calibration~\citep{zhao2021calibrate}, where they only fine-tune a small set of parameters of pre-trained models. Those works have demonstrated superior performance when training samples are not very scarce. Our experimental study, however,  show that the accuracy significantly decreases when $\# shots \leq 4$ as the limited training samples restrict the capability of learning and generalization of fine-tuning. 

There are two  considerations~when~designing~an elegant prompt learning method on PVLMs for few-shot learning. Firstly, the method should~be~generic and easily adaptable for different architectures, such as Bi-encoder structure CLIP~\citep{radford2021learning} and single encoder ViLT~\citep{kim2021vilt}. Secondly, the prompt learning method should be lightweight and competitive to or even outperforms parameter-efficient fine-tuning methods.

\newpage

In this work, we propose our model: Prompting through Prototype (\texttt{PTP}), which is a prototype-based prompt learning method on PVLMs to effectively solve the downstream few-shot image recognition tasks. Based on the observation that 1) the aligned image-text pairs have high matching scores, and 2) the similar images are close to each other in the embedding space in PVLMs, we hypothesize that similar images should use similar prompts in prompt learning. The observation 1) is because that during vision-language model pre-training, one of the pre-training objectives is image-text matching. Hence, pre-trained VL models have remarkable zero-shot performance on image-text matching. In other words, the similar images and aligned text-image paris naturally have high matching scores from PVLMs. The observation 2) will be shown during experiments. 

Intuitively, assuming training images can be coarsely divided into $K$ clusters based on the similarity between their latent embedding vectors, then each cluster can have its own textual prompt used for category name (label words) prompting. Furthermore, based on our hypothesis, we define $K$ prototype components, where each prototype component contains an image prototype and a prompt prototype. In our context, the image prototype means a point in the image latent space representing a centroid of a certain cluster. The similarity between a query image and an image prototype determines how much this query image's category prediction relies on the corresponding prompt prototype. The final prediction is the weighted summation of all the prediction scores using different prompt prototypes.   

\vspace{0.2in}
We summarize our \textbf{contributions} as follows.
\begin{itemize}
    \item We~propose~a~novel~prompt~learning~method~\texttt{PTP} on PVLMs, to overcome the drawbacks~of~task-level (manual/auto-searched prompts) and instance-level prompting. Instead~of designing a universal prompt regardless of instances~\citep{shin2020autoprompt, zhou2022learning, zhou2022cocoop}  or instance-specific prompt for each~instance~\citep{zhou2022cocoop, rao2022denseclip}, we develop a prototype-based prompting method, wherein similar query images utilizes similar prompting ways. During training, we only update parameters related to prompting while freezing the weights of PVLM to ensure a lightweight and efficient model.  
    \item We conduct extensive experiments on 7 real-world benchmarks across 2 types of PVLMs and show that our \texttt{PTP} is an effective method for the full use of the pre-trained knowledge for the downstream few-shot image recognition tasks. The absolute improvement on average accuracy compared to auto-searched prompts~\citep{zhou2022cocoop} over all experiments are around: 4\% for 1/2-shot, 5\% for 4-shot, 6\% for 8-shot, 7\% for 16-shot.
    \item  
    We made empirical analyses between prompting and fine-tuning and revealed that both methods have their advantages and limitations. In particular, a good prompt learning performance highly relies on the pre-trained knowledge stored in the pre-training. A prompt learning method will have difficulty triggering the correct answers, if the PVLM itself lacks such visual or textual knowledge. 
    Through detailed hyper-parameter analysis, we show how to choose the number of prototypes based on performance and parameter-efficiency. We also show the importance of our novel regularizers for learning the image prototypes. 
\end{itemize}

\newpage

\section{Related Work}

\subsection{Pretrained Vision-and-Language Models}
Recently, many vision-language models are proposed. The large-scale pre-training allows PVLMs to zero-shot transfer to various downstream classification tasks. They can be coarsely divided into two groups based on their architecture: the bi-encoder model~\citep{radford2021learning, jia2021scaling}, and the single-encoder model~\citep{kim2021vilt, lu2019vilbert}. 
Bi-encoder model, such as CLIP~\citep{radford2021learning} and ALIGN~\citep{jia2021scaling}, consists of two encoders, one for images and the other for text. This work uses CLIP as a representative for the bi-encoder model, which has remarkable zero-shot performance on image-text retrieval.. By default, CLIP uses ``a photo of [CLASS]'' on the text side for image recognition tasks. 

Single-encoder model, such as ViLBERT~\citep{lu2019vilbert}, ViLT~\citep{kim2021vilt}, etc., concatenates the object features from the image and word features from the sentence into a long sequence. So the two modalities interact with each other in self-attention layers. This work uses ViLT as a representative for single-encoder models.

\subsection{Few-shot Learning}

\noindent \textbf{Parameter-Efficient Fine-tuning.}\ Parameter-efficient fine-tuning methods mainly include: i) Adapters~\citep{houlsby2019parameter, gao2021clip, zhang2021tip}, where neural networks layers are inserted between the feed-forward portion of the Transformer architecture; ii) BitFit~\citep{zaken2022bitfit, logan2022cutting}, where they only update the bias terms inside the Transformer; iii) Calibration~\citep{zhao2021calibrate}, where they learn an affine transformation on top of the logits output from the Transformer; iv) Linear probe~\citep{tian2020rethinking}, where a linear classifier is trained on top of pre-trained models' features. 

\vspace{10pt}
\noindent \textbf{Prompt Learning Methods.}\ Recently, multiple prompt learning works on PVLM are proposed~\citep{jin2022good, zhou2022learning, tsimpoukelli2021multimodal, liang2022modular, rao2022denseclip}.~\citet{jin2022good} first pre-trained a prompt-aware vision language model, then transferred to downstream tasks, such as VQA, with the help of hand-crafted prompts.~\citet{zhou2022learning} learned universal soft prompts for solving downstream few-shot image classification tasks.~\citet{tsimpoukelli2021multimodal} developed an image to text generation model, with a dynamic prefix to control the generation.~\citet{liang2022modular} learned soft prompts to align the different modalities.~\citet{rao2022denseclip} learned instance-aware prompts for dense prediction. 

In this work,  we focus on designing an efficient and effective prompt learning method on PVLMs for downstream few-shot image classification tasks. We leverage prototype-based prompting. Our image prototypes have a similar concept and usage ``this looks like that'' in previous works~\citep{li2018deep, chen2019looks}, where they learn and utilize prototypes to make interpretable predictions.

\section{Methodology}
\subsection{Problem Setup}
We define a few-shot image recognition training dataset as $D=\{(x_i, y_i, c_i)\}_{i=1}^{N}$, where $x_i$ is the image input, $y_i$ is corresponding discrete label, $c_i$ is corresponding category name, e.g., ``faces'', ``Hummer SUV''. We define the candidate pool of category names as $\mathcal{C}=\{c_j\}_{j=1}^{C}$, where $C$ is total number of different categories. Given a pre-trained vision-language model (PVLM) and a few-shot training dataset $D$, our task aims at solving the downstream few-shot image classification task via prompt learning paradigm. 

\begin{figure*}[!tb]
\centering
\includegraphics[width=1\textwidth]{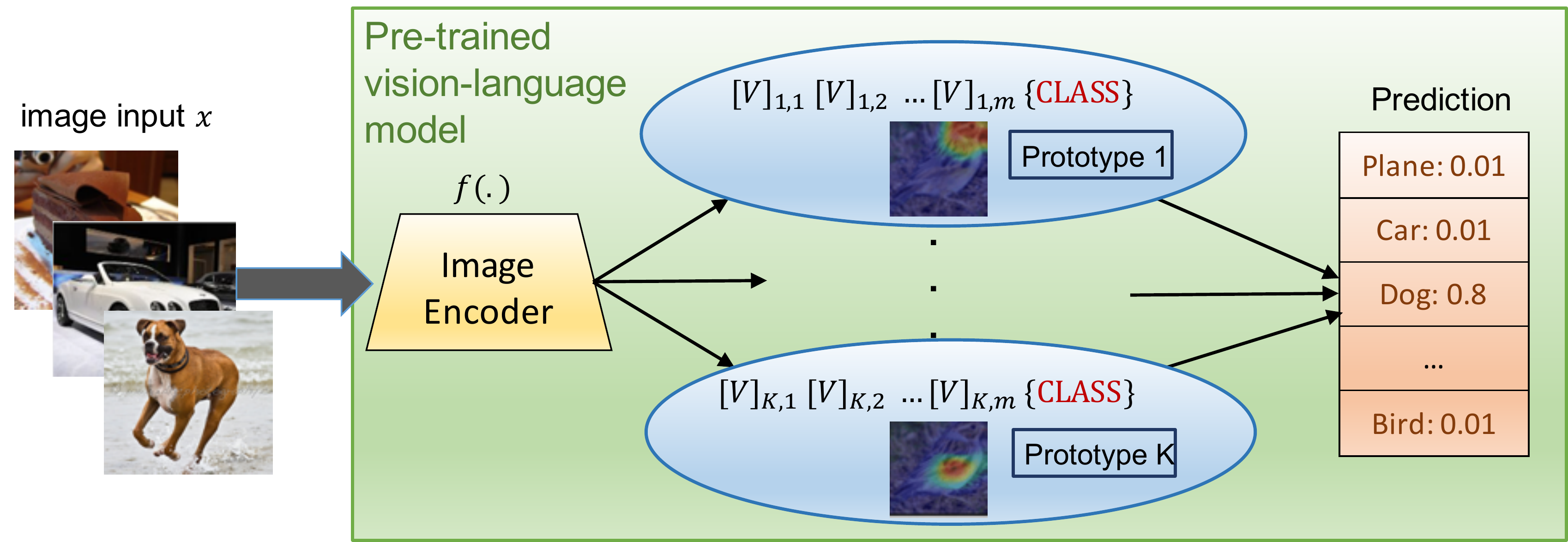} %
\caption{The overall architecture of our model \texttt{PTP}. \texttt{PTP} mainly consists of: i) an image encoder, which is a part of PVLM; ii) $K$ prototype components, where each component consists of an image prototype $\mathcal{P}_k$ and a prompt prototype $\mathcal{T}_k$; iii) a fixed PVLM. During training, we learn the lightweight parameters related to prompting, i.e., image prototypes and prompt prototypes, and keep the pre-trained vision-language model \textbf{frozen}. } 
\label{fig:architecture}
\end{figure*}
\subsection{Model Architecture}
The overall architecture of our model is shown in Figure~\ref{fig:architecture}. Our \texttt{PTP} model consists of three major parts: i) a pre-trained and fixed image encoder $f(.)$, which is a part of a PVLM; ii) $K$ prototype components, where each prototype component consists of an image prototype $\mathcal{P}_k$ and a prompt prototype $\mathcal{T}_k$, $k \in \{1,2,\cdots,K\}$; iii) a fixed PVLM, which takes an image, prompted category name as input and outputs their matching score. 

\subsubsection{Pre-trained Image Encoder}

The image encoder $f(.)$ takes image $x$ as input and outputs the image latent representation $f(x) \in \mathbb{R}^d$. Bi-encoder PVLMs, such as CLIP, incorporate two encoders, one for image and the other for text. So, for bi-encoder PVLMs, we can directly utilize its pre-trained image encoder. 

While, single-encoder PVLMs, such as ViLT, do not have a standalone image encoder by default. For single-encoder PVLMs, we calculate their image encoding by putting the query image and an empty text as the input of an single-encoder PVLM:
\begin{align*}
    f(x) = \textrm{PVLM}_{\textrm{pooler}}(x, [\textrm{CLS}][\textrm{SEP}]),
\end{align*}
where PVLM is a single-encoder model, $f(x)$ is the pooler output of the PVLM, the text side only contains special tokens [CLS] and [SEP], which are pre-defined in the PVLM vocabulary. 

Since $f(.)$ is a part of a PVLM, through large-scale pre-training, it has the ability to map similar images into close latent vectors. During training, we keep $f(.)$ frozen. Then, the encoded image representation $f(x)$ is used to calculate the similarity score with each image prototype $\mathcal{P}_k$: 
\begin{align*}
    \textrm{sim}(x, \mathcal{P}_k) = \frac{\textrm{exp}(f(x)^T \cdot \mathcal{P}_k)}{\sum_{i=1}^K \textrm{exp}(f(x)^T \cdot \mathcal{P}_i)}
\end{align*}

\newpage

\subsubsection{Prototype Component}

In our few-shot setting, we define $K$ prototype components, where $K$ is much smaller than the total number of class $C$. Each prototype component consists of an image prototype and a prompt prototype. In total, we have $K$ image prototypes and $K$ prompt prototypes. 

\vspace{0.1in}
\noindent \textbf{Image Prototype.}\  The image prototypes are defined in the image latent space, $\mathcal{P}_k \in \mathbb{R}^d$. During training, we define two regularizers to encourage the learned image prototype vectors to correspond to meaningful points in the latent space. In our setting, we hypothesize that image prototype vectors are the centroids of certain clusters of training image data. The two regularizers are formulated as follows:
\begin{align*}
    R_1(\mathcal{P}_1, ..., \mathcal{P}_K, D) &= \frac{1}{K} \sum_{j=1}^K \min_{i \in [1, N]} || \mathcal{P}_j-f(x_i)||_2^2, \\
    R_2(\mathcal{P}_1, ..., \mathcal{P}_K, D) &= \frac{1}{N} \sum_{i=1}^N \min_{j \in [1, K]} || f(x_i)- \mathcal{P}_j||_2^2.
\end{align*}
Here both terms are averages of minimum squared distances. The minimization of $R_1$ would require each image prototype to be close to at least one of the training examples in the latent space. $R_1$ will push each image prototype to be a centroid of a certain cluster. The minimization of $R_2$ would require every encoded training example to be close to one of the image prototype vectors. This means that $R_2$ will cluster the training examples around image prototypes in the latent space. We relax $R_1$ minimization to be over only the random minibatch during stochastic gradient descent. 

\vspace{0.1in}
\noindent \textbf{Prompt Prototype.}\ Each image prototype has its corresponding prompt prototype. Our prompt prototype is defined in the continuous space in the form of:
\begin{align*}
    \mathcal{T}_k = [V]_{k,1} \hspace{0.1cm} [V]_{k,2} \hspace{0.1cm}... \hspace{0.1cm} [V]_{k,m} \{\textrm{CLASS}\},
\end{align*}
where each $[V]_{k,i} \in \mathbb{R}^d$ is a dense vector with the same dimension as the PVLM's input embedding, and the number of $[V]$ vectors is set to a pre-defined number $m$. Compared with using discrete vocabulary tokens as prompts, continuous prompts can be viewed as adding new tokens into pre-defined vocabulary, and each $[V]_{k,i}$ is the word embedding of a new token. In our \texttt{PTP}, we define $K$ prompt prototypes, and each prompt prototype contains $m$ new tokens. So totally, we add $m*K$ new tokens into pre-defined vocabularies, and we will only update the word embedding of new tokens. The values for $[V]_{k,i}$ can be randomly initialized, and updated through gradient descent. The $\{\textrm{CLASS}\}$ represents any candidate category name.

For bi-encoder PVLMs, such as CLIP, our prompts only affect its textual input, keeping the image side unchanged. While, for single-encoder PVLMs, such as ViLT, since it concatenates and fuses image and text information, our prompts have the ability to change the image-text pair input.

\subsubsection{PVLM for Classification}

PVLM model takes image-text pair as input and outputs their matching score. An image classification problem can be converted into an image-text pair matching problem, where the image side is our query image, the text side is our category name with prompting. Finally, we can get the matching scores on all candidate categories.  Specifically, given a query image $x$, a prompt $\mathcal{T}_k$, a concrete category name $c$, we have its matching score under prompt $\mathcal{T}_k$ equals to:
\begin{align*}
    \textrm{match}_{\mathcal{T}_k}(x, c)= \textrm{PVLM}(x, \mathcal{T}_k(c)),
\end{align*}
where $\mathcal{T}_k(c)$ means we replace $\{\textrm{CLASS}\}$ in prompt $\mathcal{T}_k$ using a category $c$. 

\vspace{5pt}
\noindent \textbf{Bi-encoder PVLM.}\ For a bi-encoder PVLM, such as CLIP, calculating the matching score can be further decomposed into:
\begin{align*}
     \textrm{match}_{\mathcal{T}_k}(x, c) = f(x)^T \cdot g(\mathcal{T}_k(c)),
\end{align*}
where $g(.)$ represents the text encoder. Also, the prediction scores over all the candidate categories under the prompt $\mathcal{T}_k$ are then computed as:
\begin{align*}
    \textrm{Prob}_{\mathcal{T}_k}(y=i|x)=\frac{\exp(\textrm{match}_{\mathcal{T}_k}(x, c_i)/\tau)}{\sum_{j=1}^C \exp(\textrm{match}_{\mathcal{T}_k}(x, c_j)/\tau)},
\end{align*}
where $\tau$ is a temperature parameter learned by CLIP. 

\vspace{0.1in}
\noindent \textbf{Single-encoder PVLM.}\ For single-encoder PVLM, such as ViLT, the prediction score for each category under the prompt $\mathcal{T}_k$ is then computed as:
\begin{align*}
    \textrm{Prob}_{\mathcal{T}_k}(y=i|x)= \textrm{Sigmoid}(\textrm{match}_{\mathcal{T}_k}(x, c_i)),
\end{align*}
Here, we use different functions to calculate prediction scores for CLIP and ViLT, respectively. During pre-training, CLIP utilizes contrastive loss for the image-text matching objective, while ViLT utilizes binary cross-entropy loss. So, we follow their convention in their pre-training. 

Considering all the $K$ prompts together, our final prediction score on each category equals the weighted summation of prediction scores under different prompts, which is formulated in the following:
\begin{align*}
    \textrm{Prob}(x, c) = \sum_{k \in [1,K]}\textrm{sim}(x, \mathcal{P}_k) *  \textrm{Prob}_{\mathcal{T}_k}(x, c),
\end{align*}
At inference time, we will choose the category with the highest matching score as our final classification result. We can see that if input image $x$ is similar to an image prototype $\mathcal{P}_k$ on the latent space, then its classification result is more dependent on the prompt $\mathcal{T}_k$.

\subsection{Objective Function}

Our first cost function reflects the classification error. For bi-encoder PVLM, we compute the cross-entropy loss for penalizing the misclassification. The cross-entropy loss on the training data $D$ is denoted by $E$, and is given by:
\begin{align*}
    E(\theta, D) = \frac{1}{N} \sum_{i=1}^N \sum_{j=1}^C -\mathds{1}[y_i=j]\log(\textrm{Prob}(x_i, c_j)), 
\end{align*}
For single-encoder PVLM, we compute the binary cross-entropy loss for penalizing the misclassification as follows:
\begin{align*}
    E(\theta, D) &= -\frac{1}{N} \sum_{i=1}^N \sum_{j=1}^C [y_j \cdot \log(\textrm{Prob}(x_i, c_j)) +(1-y_j) \cdot \log(\textrm{Prob}(x_i, c_j))]
\end{align*}
where $\theta$ represents all the parameters related to prompting, i.e., $\theta=\{ \mathcal{P}_1, ...., \mathcal{P}_K, \mathcal{T}_1, ...., \mathcal{T}_K\}$. 

Putting everything together, the cost function, denoted by $L$, on the training data $D$, is given by:
\begin{align*}
    L(\theta, D) &= E(\theta, D)+ \lambda \cdot [R_1(\mathcal{P}_1, ..., \mathcal{P}_K, D)+ R_2(\mathcal{P}_1, ..., \mathcal{P}_K, D)], 
\end{align*}
where $\lambda \in [0, 1]$ is a hyper-parameter that adjusts the weight of the regularizers. Our model \texttt{PTP} is trained end-to-end.

\section{Experiment}

\subsection{Datasets and Experiment Setting}

We evaluate \texttt{PTP} on 7 publicly available image classification datasets: 1) Caltech101, which contains 100 classes, such as ``faces'', ``leopards'', ``laptop'', etc., and contains 2,465 test data. 2) StanfordCars~\citep{krause20133d}, which contains 196 classes on cars, such as ``2000 AM General Hummer SUV'', ``2012 Acura RL Sedan'', etc., and contains 8,041 test data. 3) OxfordPets~\citep{parkhi2012cats}, which contains 37 classes on pets, such as ``beagle'', ``chihuahua'', etc., and contains 3,669 test data. 4) UCF101 dataset~\citep{soomro2012ucf101}, where the middle frame of each video is used as image input; it contains 101 classes on people activity, such as ``baby crawling'', ``biking'', ``blow dry hair'', etc., and contains 3,783 test data. 5) Food101~\citep{bossard2014food}, which contains 101 classes on food, such as ``apple pie'', ``crab cakes'', etc., and contains 30,300 test data.6) SUN397~\citep{xiao2010sun}, which contains 397 classes, such as ``barn'', ``campus'', etc., and contains 19,850 test data.  7) FGVCAircraft~\citep{maji2013fine}, which contains 100 classes on aircraft models, such as ``Boeing 717'', ``DC-10'', ``EMB-120'', etc., and contains 3,333 test data. 

For PVLM models, we consider bi-encoder CLIP~\citep{radford2021learning} and single-encoder ViLT~\citep{kim2021vilt}. For both CLIP and ViLT, we use ViT-B/32 as image encoder backbone. Compared with ResNet based backbone, ViT-B/32 is more powerful on image encoding and has better zero-shot transfer ability~\citep{radford2021learning}. The parameter size of \texttt{PTP} equals to $m* d* K+ K* d= (m+1)*d*K$. We setup prototype number $K$ considering both the performance and the parameter-efficiency. Table~\ref{tab:hyper-parameter} shows the hyper-parameter settings of \texttt{PTP}. Since a bi-encoder is much more computationally efficient than a single-encoder, for ViLT, we set a smaller prompt token number $m$.

\begin{table}[ht]
    \centering
    \caption{Hyper-parameter setting of our model \texttt{PTP}.}
    \begin{tabular}{cc}
    \toprule
    Hyper-parameter & Value \\
    \midrule
    $m$ for CLIP & 16 \\
    $m$ for ViLT & 5 \\ 
    $\lambda$ & 1.0 \\
    optimizer & AdamW \\
    learning rate & 3e-3 \\
    warm-up rate & 0.1 \\
    batch size & 32 \\
    image encoder & ViT-B/32 \\
    \bottomrule
    \end{tabular}
    \label{tab:hyper-parameter}
\end{table}

We follow the few-shot evaluation protocol adopted in previous work~\citep{radford2021learning, zhou2022learning}, using 1,2,4,8 and 16 shots for training, respectively, and deploying models in the full test sets. Here, the $n$ shots means $n$ number of labeled training examples per class. In our few-shot setting, the hold-out validation dataset is unavailable. The average accuracy results on the test set over three runs are reported for comparison. More specifically, For experiments on CLIP, the maximum epoch is set to 200 for 16/8 shots, 100 for 4/2/1/ shots. For experiments on ViLT, the maximum epoch is set to 1000 for 16/8 shots, 500 for 4/2/1/ shots. For baseline CoCoOp~\citep{zhou2022cocoop}, since it is easily overfitting on the few-shot dataset, the maximum epoch is set to 50 for all CoCoOp experiments, except for StanfordCars and SUN397 datasets, which we set to 10. The average accuracy results on the test set over three runs are reported for comparison. 

\subsection{Baselines}

First, we compare with zero-shot baselines: 1) Manual-crafted prompt (MCP)~\citep{radford2021learning}; 2) Vision matching (VM).  Secondly, we compare with parameter-efficient fine-tuning methods: 3) Linear probe (LP)~\citep{tian2020rethinking}; 4) Bitfit fine-tuning~\citep{zaken2022bitfit}. Thirdly, we compare with other prompt learning baselines: 5) Soft prompt (SP)~\citep{zhong2021factual, zhou2022learning}; 6) CoCoOp~\citep{zhou2022cocoop}. 

Table~\ref{tab:baseline_detail} gives the details of each baseline, including their parameter size, where $C$ is the total class number, $d$ is the latent space dimension of a PVLM, $m$ is the number of prompt vector $[V]$,  $h$ is the middle layer size of the neural nets. In practice, the possible value of $h$ could be 32, $m$ could be 16. For CLIP and ViLT, $d$ equals 512. 

\begin{table*}[!htb]
    \centering
    \caption{Details of comparison baselines.}
    \begin{tabular}{p{5cm}p{7cm}c}
    \toprule
    Baseline & Description & Parameter size \\
    \midrule
    Manual-crafted prompt (MCP)~\citep{radford2021learning} & It uses the default prompt ``a photo of a \{CLASS\}''; for the UCF101 dataset, we use the prompt ``a photo of a person doing \{CLASS\}'' & 0 \\
    \midrule
    Vision matching (VM) & For CLIP, given n-shot training data per class, we calculate the mean of the image latent vectors in each category as the representation for each class. Then we match the query images with each class representation in the image latent space. & 0 \\
    \midrule
    Linear probe (LP)~\citep{tian2020rethinking} & It is a linear classifier on top of CLIP's image encoder. & $d* C$ \\
    \midrule
    Bitfit fine-tuning~\citep{zaken2022bitfit} & It only updates the bias terms in PVLM, also it uses ``a photo of a \{CLASS\}'' at text side. & $\approx10^3$ \\
    \midrule
    Soft prompt (SP)~\citep{zhong2021factual, zhou2022learning} & it learns a universal prompt with a template $[V]_1 \hspace{0.1cm}... [V]_m \hspace{0.1cm} \{\textrm{CLASS}\}.$ & $m* d$ \\
    \midrule
    CoCoOp~\citep{zhou2022cocoop} & It is an instance-level prompt learning method designed on CLIP. It maintains two-layer neural nets. & $(2h+m)* d$\\
    \bottomrule
    \end{tabular}
    \label{tab:baseline_detail}
\end{table*}

\subsection{Image Recognition Result using CLIP}
We set prototype number $K=3$ for SUN397; $K=5$ for Caltech101, UCF101, Food101, and FGVCAircraft; $K=7$ for StanfordCars, and OxfordPets. We report the comparison results on 7 datasets using CLIP from Table~\ref{tab:clip_caltech} to Table~\ref{tab:clip_aircraft}, respectively. 

From Tables~\ref{tab:clip_caltech}-\ref{tab:clip_aircraft}, firstly, we see that the zero-shot method manual-crafted prompt (MCP) gets the overall acceptable classification accuracy. Especially in Table~\ref{tab:clip_cars} at 1-shot, MCP gets the best performance. Secondly, we see that with the training data increasing (i.e., from 1-shot to 16-shot), the performance of vision matching (VM) increases. On some datasets, at 8/16 shots, VM method can even outperform the MCP method, such as in Tables~\ref{tab:clip_caltech} and Table~\ref{tab:clip_ucf101}.

\begin{table}[ht]
    \centering
    \caption{Accuracy on Caltech101 dataset using CLIP.}
    \begin{tabular}{cccccc}
    \toprule
    Method & 1 & 2 & 4 & 8 & 16 \\
    \midrule
    MCP & 88.52 & 88.52 & 88.52 & 88.52 & 88.52 \\
    VM & 74.72 & 81.50 & 86.09 & 89.09 & 90.02 \\
    \midrule
    LP & 77.32 & 86.41 & 90.51 & 94.36 & 95.21 \\
    Bitfit & 88.60 & 90.02 & 92.25 & 94.44 & 95.17 \\
    \midrule
    SP & 89.12 & 91.44 & 92.57 & 92.86 & 93.34 \\
    CoCoOp & 90.87 & 89.94 & 91.85 & 92.70 & 93.75 \\
    PTP & \textbf{91.93} & \textbf{93.31} & \textbf{94.81} & \textbf{95.29} & \textbf{96.11} \\
    \bottomrule
    \end{tabular}
    \label{tab:clip_caltech}
\end{table}

\begin{table}[ht]
    \centering
    \caption{Accuracy on StanfordCars dataset using CLIP.}
    \begin{tabular}{cccccc}
    \toprule
    Method & 1 & 2 & 4 & 8 & 16 \\
    \midrule
    MCP & \textbf{60.30} & 60.30 & 60.30 & 60.30 & 60.30 \\
    VM & 28.06 & 34.27 & 41.29 & 49.01 & 53.65 \\
    \midrule
    LP & 29.95 & 42.59 & 56.85 & 66.35 & 74.11 \\
    Bitfit & 42.59 & 52.23 & 58.68 & 63.08 & {69.25} \\
    \midrule
    SP & 54.70 & 56.74 & 58.02 & 59.88 & 60.58 \\
    CoCoOp & 48.44 & 59.06 & 62.16 & 63.26 & 65.02 \\
    PTP & 59.72 & \textbf{61.67} & \textbf{66.10} & \textbf{71.99} & \textbf{75.71} \\
    \bottomrule
    \end{tabular}
    \label{tab:clip_cars}
\end{table}

\begin{table}[ht]
    \centering
    \caption{Accuracy on Oxford Pets dataset using CLIP.}
    \begin{tabular}{cccccc}
    \toprule
    Method & 1 & 2 & 4 & 8 & 16 \\
    \midrule
    MCP & 79.97 & 79.97 & 79.97 & 79.97 & 79.97 \\
    VM & 36.82 & 42.71 & 50.80 & 57.51 & 64.51 \\
    \midrule
    LP & 38.51 & 55.93 & 66.37 & 75.66 & 81.66 \\
    Bitfit & 76.15 & 81.74 & 87.44 & 87.82 & 88.44 \\
    \midrule
    SP & 77.84 & 82.12 & 86.70 & 87.84 & 89.45 \\
    CoCoOp & 81.28 & 82.69 & 81.25 & 79.94 & 87.60 \\
    PTP & \textbf{83.40} & \textbf{86.56} & \textbf{89.39} & \textbf{90.97} & \textbf{91.66} \\
    \bottomrule
    \end{tabular}
    \label{tab:clip_pets}
\end{table}

\begin{table}[ht]
    \centering
    \caption{Accuracy on UCF101 dataset using CLIP.}
    \begin{tabular}{cccccc}
    \toprule
    Method & 1 & 2 & 4 & 8 & 16 \\
    \midrule
    MCP & 62.09 & 62.09 & 62.09 & 62.09 & 62.09 \\
    VM & 46.54 & 56.2 & 60.68 & 64.24 & 67.29 \\
    \midrule
    LP & 47.80 & 63.21 & 70.69 & 75.62 & 79.85 \\
    Bitfit & 61.71 & 65.78 & 71.7 & 74.28 & 78.74 \\
    \midrule
    SP & 61.85 & 64.10 & 66.72 & 69.31 & 71.38 \\
    CoCoOp & 58.08 & 60.80 & 62.38 & 72.67 & 74.20 \\
    PTP & \textbf{65.98} & \textbf{68.99} & \textbf{72.93} & \textbf{77.66} & \textbf{80.02} \\
    \bottomrule
    \end{tabular}
    \label{tab:clip_ucf101}
\end{table}

\begin{table}[ht]
    \centering
    \caption{Accuracy on Food101 dataset using CLIP.}
    \begin{tabular}{cccccc}
    \toprule
    Method & 1 & 2 & 4 & 8 & 16 \\
    \midrule
    MCP &  82.34 & 82.34 & 82.34 & 82.34 & 82.34 \\
    VM & 38.07 &  45.78 & 55.24 & 63.18 & 68.04 \\
    \midrule
    LP & 41.69 & 55.66 & 67.99 & 75.49 & 78.92 \\
    Bitfit & 68.86 & 71.26 & 76.76 & 75.82 & 76.85 \\
    \midrule
    SP & 81.86 & 82.19 & 82.90 & 83.21 & 83.49 \\
    CoCoOp & 67.62 & 69.18 & 69.90 & 72.85 & 75.39 \\
    PTP & \textbf{83.33} & \textbf{83.63} & \textbf{84.26} & \textbf{84.44} & \textbf{84.91} \\
    \bottomrule
    \end{tabular}
    \label{tab:clip_food101}
\end{table}

\begin{table}[ht]
    \centering
    \caption{Accuracy on SUN397 dataset using CLIP.}
    \begin{tabular}{cccccc}
    \toprule
    Method & 1 & 2 & 4 & 8 & 16 \\
    \midrule
    MCP & 60.16 & 60.16 & 60.16 & 60.16 & 60.16 \\
    VM & 34.85 & 44.42 & 50.52 & 55.69 & 59.34 \\
    \midrule
    LP &  36.88 & {51.88} & {61.72} & 66.21 & {70.19} \\
    Bitfit & 58.20 & 62.39 & 64.52 & 64.53 & 67.23 \\
    \midrule
    SP & 62.32 & 63.94 & 65.13 & 66.55 & 67.32 \\
    CoCoOp & 62.85 & 64.80 & 66.19 & 66.45 & 67.57 \\
    PTP &  \textbf{63.51} & \textbf{65.72} & \textbf{68.76} & \textbf{70.93} & \textbf{72.42} \\
    \bottomrule
    \end{tabular}
    \label{tab:clip_sun397}
\end{table}

\begin{table}[ht]
    \centering
    \caption{Accuracy on FGVCAircraft dataset using CLIP.}
    \begin{tabular}{cccccc}
    \toprule
    Method & 1 & 2 & 4 & 8 & 16 \\
    \midrule
    MCP & 17.52 & 17.52 & 17.52 & 17.52 & 17.52 \\
    VM & 17.52 & 18.96 & 21.72 & 24.15 & 25.41 \\
    \midrule
    LP & 17.22 & \textbf{20.73} & \textbf{27.24} & \textbf{33.33} & 38.61 \\
    Bitfit & 14.19 & 17.07 & 23.16 & 33.15 & \textbf{40.56} \\
    \midrule
    SP & 18.09 & 17.07 & 16.98 & 19.77 & 22.23  \\
    CoCoOp & 14.73 & 16.56 & 18.90 & 25.53 & 29.22\\
    PTP & \textbf{19.62} & 19.83 & 23.7 & 26.86 & 33.21 \\
    \bottomrule
    \end{tabular}
    \label{tab:clip_aircraft}
\end{table}

\newpage\clearpage

In baseline vision matching (VM), given n-shot training data per class, we first calculate the mean of the image latent vectors in each category as the representation for each class. Then we match the query images with each class representation in the image latent space without fine-tuning. This VM baseline achieves overall good performance on the 16-shot scenarios, hence it proves our hypothesis that similar images are close to each other. 

For parameter-efficient fine-tuning methods linear probe (LP) and Bitfit, overall, we see that when there is less training data (i.e., 1/2-shot), LP and Bitfit can not perform as well as other baselines, sometimes even much worse than the zero-shot baselines, such as at 1/2-shot in Tables~\ref{tab:clip_cars},~\ref{tab:clip_pets} and~\ref{tab:clip_food101}. In Table~\ref{tab:clip_cars} at 1-shot, the MCP gets the accuracy of 60.3, while LP only gets an accuracy of 29.95, where we can see a huge performance gap between them. While, with the training data increasing (i.e., 8/16-shot), LP and Bitfit become very strong baselines. 

Overall, prompt learning baselines soft prompt (SP) and CoCoOp perform better on 1/2-shots, compared with LP and Bitfit. CoCoOp is not as effective as SP in many cases.  It is because CoCoOp is not as lightweight as SP or our \texttt{PTP}, which makes it easily over-fitting in few-shot scenarios. 

We observe that except in Table~\ref{tab:clip_aircraft}, our \texttt{PTP} outperforms all zero-shot baselines, prompt learning baselines and parameter-efficient fine-tuning baselines. Even at 1-shot, \texttt{PTP} has the ability to achieve superior performance. With the training data increasing, our \texttt{PTP} still can outperform the very strong fine-tuning baselines.

Now, lets take a close look at Table~\ref{tab:clip_aircraft}. From Table~\ref{tab:clip_aircraft}, we observe that VM outperforms MCP, which means for the FGVCAircraft dataset, image-image matching works better than image-text matching. We see that \texttt{PTP} achieves the best performance on 1-shot. However, on 2/4/8/16-shot, LP and Bitfit achieve the overall best performance. Through detailed analysis, we find that many category names provided by the FGVCAircraft dataset do not have semantic meaning, such as ``707-320'', ``A321'' and ``BAE-125'', etc. These semantic meaningless category names make prompt learning methods inferior. But, we can see that our \texttt{PTP} still can get the best performance over other prompt learning methods, given these challenging category names. On the other hand, LP does not reply on category names, and Bitfit can update the word embedding and learn the semantics of ``707-320'', ``A321'' and ``BAE-125'', etc., through fine-tuning.

\subsection{Image Recognition Result using ViLT}

ViLT itself is not as powerful as CLIP on image-text matching, because ViLT is pre-trained using smaller data, which results in ViLT cannot zero-shot or few-shot transfer to very fine-grained classification tasks. So, for ViLT, we only conduct experiments on three datasets: Caltech101, UCF101, and Food101. For all three datasets, we set $K=5$. 

\begin{table}[ht]
    \centering
    \caption{Accuracy on Caltech101 dataset using ViLT.}
    \begin{tabular}{cccccc}
    \toprule
    Method & 1 & 2 & 4 & 8 & 16 \\
    \midrule
    MCP & 58.58 & 58.58 & 58.58 & 58.58 & 58.58 \\
    \midrule
    Bitfit & {49.04} & {49.70} & 54.92 & 72.53 & 75.21 \\
    \midrule
    SP & 54.93 & 64.30 & 68.32 & 74.32 & 77.28\\
    PTP & \textbf{67.42} & \textbf{69.53} & \textbf{71.40} & \textbf{77.12} & \textbf{80.73}\\
    \bottomrule
    \end{tabular}
    \label{tab:vilt_caltech}
\end{table}

\begin{table}[ht]
    \centering
    \caption{Accuracy on UCF101 dataset using ViLT.}
    \begin{tabular}{cccccc}
    \toprule
    Method & 1 & 2 & 4 & 8 & 16 \\
    \midrule
    MCP &  29.37 & 29.37 & 29.37 & 29.37 & 29.37 \\
    \midrule
    Bitfit & 31.95 & 33.17 & {41.79} & \textbf{55.16} & \textbf{63.39} \\
    \midrule
    SP & 31.64 & 33.25 & 37.11 & 44.65 & 54.93  \\
    PTP & \textbf{34.44} & \textbf{36.37} & \textbf{41.98} & 53.90 & 58.84 \\
    \bottomrule
    \end{tabular}
    \label{tab:vilt_ucf101}
\end{table}

\begin{table}[ht]
    \centering
    \caption{Accuracy on Food101 dataset using ViLT.}
    \begin{tabular}{cccccc}
    \toprule
    Method & 1 & 2 & 4 & 8 & 16 \\
    \midrule
    MCP & 15.41 & 15.41 & 15.41 & 15.41 & 15.41 \\
    \midrule
    Bitfit & 18.93 & 20.02 & 22.60  & \textbf{27.39} & \textbf{35.57} \\
    \midrule
    SP & 16.67 & 16.95 & 20.53 & 22.79 & 24.78  \\
    PTP & \textbf{19.01} & \textbf{20.50} & \textbf{22.67} & 27.16 & {32.06} \\
    \bottomrule
    \end{tabular}
    \label{tab:vilt_food101}
\end{table}

We report the comparison results on these three datasets using ViLT in Table~\ref{tab:vilt_caltech},~\ref{tab:vilt_ucf101} and~\ref{tab:vilt_food101}, respectively. In Table~\ref{tab:vilt_caltech}, we see \texttt{PTP} significantly outperforms zero-shot manual prompt method, Bitfit fine-tuning method, and prompt learning baseline SP. In Table~\ref{tab:vilt_ucf101} and Table~\ref{tab:vilt_food101}, we see \texttt{PTP} outperforms all other baselines at 1/2/4-shot scenarios.

However, in Table~\ref{tab:vilt_ucf101} and~\ref{tab:vilt_food101}, at 8/16-shots, Bitfit fine-tuning method becomes better than \texttt{PTP}. With more training data, Bitfit fine-tuning method updates the parameters of PVLM towards optimal.  But, we see \texttt{PTP} still can outperform prompt learning baseline SP on all the cases. 

\subsection{Discussion and Analysis}

\begin{figure*}
\mbox{\hspace{-0.3in}
     \begin{subfigure}{0.35\textwidth}
         \centering
         \includegraphics[width=\textwidth]{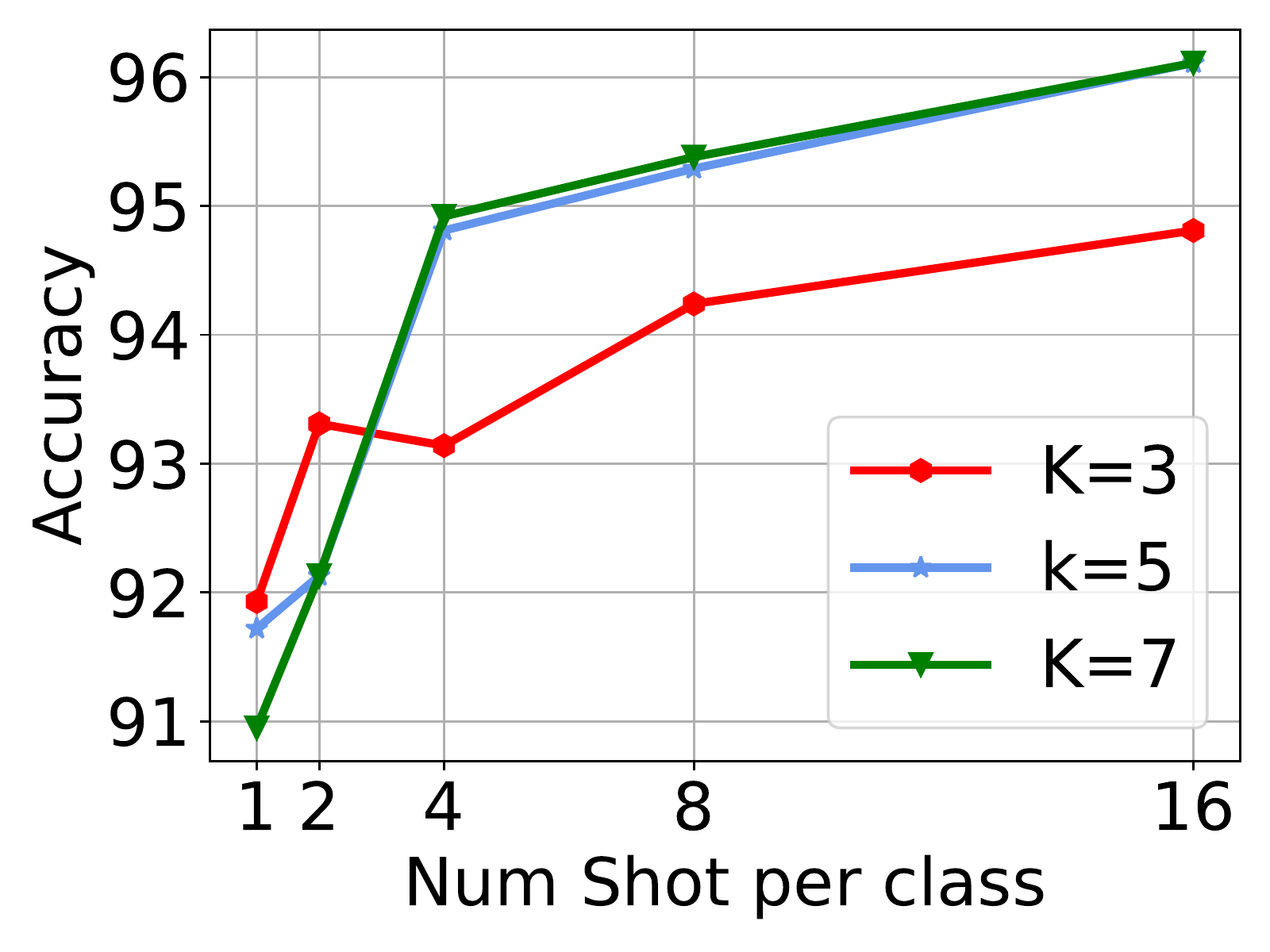}
         \caption{Caltech101 dataset.}
         \label{fig:1}
     \end{subfigure}
     \hspace{-0.1in}
     \begin{subfigure}{0.35\textwidth}
         \centering
         \includegraphics[width=\textwidth]{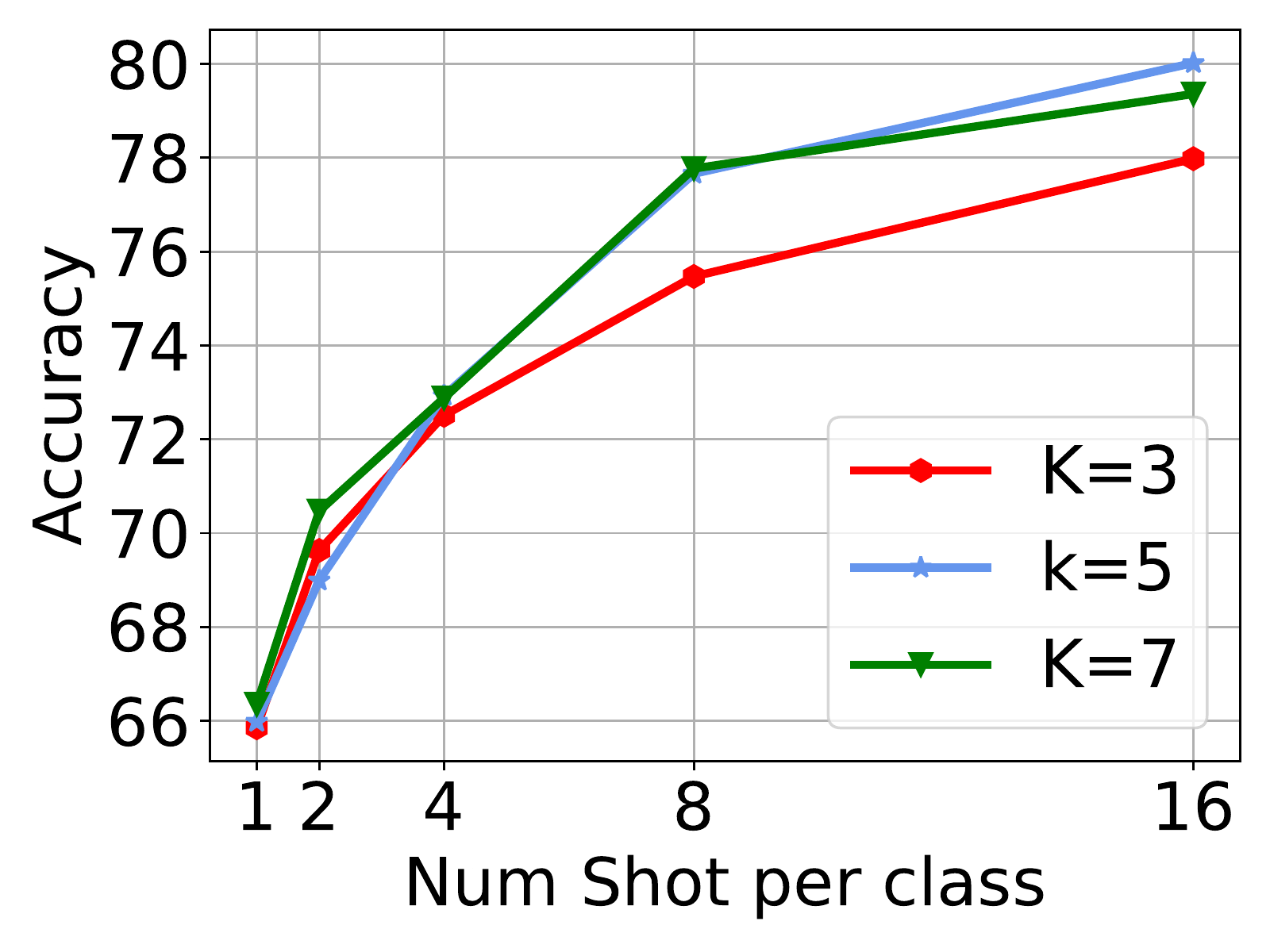}
         \caption{UCF101 dataset.}
         \label{fig:2}
     \end{subfigure}
     \hspace{-0.1in}
     \begin{subfigure}{0.35\textwidth}
         \centering
         \includegraphics[width=\textwidth]{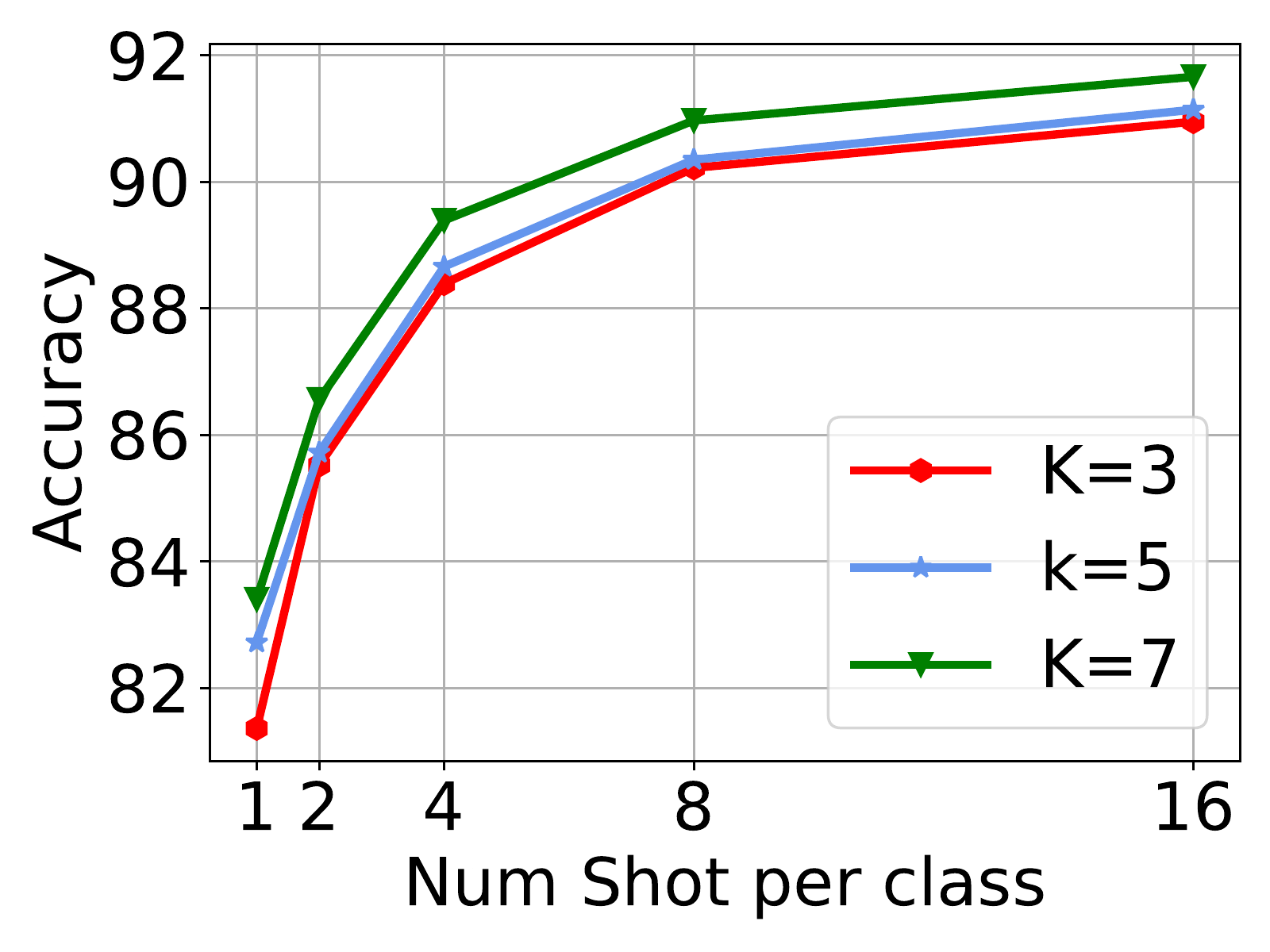}
         \caption{OxfordPets dataset.}
         \label{fig:3}
     \end{subfigure}
}
        \caption{Prototype number $K$ analysis on three dataset using CLIP.}
        \label{fig:K_analysis}
\end{figure*}

\subsubsection{Analysis on Prototype Number $K$}
First, we analyze the model performance under different prototype numbers $K$. We conduct an analysis using CLIP. We set $K=3, 5, 7$, respectively, and test on the three datasets: Caltech101, UCF101 and OxfordPets. The results are reported in Figure~\ref{fig:K_analysis}.

We can see that at 1/2-shot scenarios, higher $K$ does not necessarily lead to higher performance, such as in Figure ~\ref{fig:K_analysis} (a), the best performance at 1-shot comes from $K=3$. At 4, 8, and 16-shot, we see the general trend is that higher $K$ leads to higher performance. However, from Figure ~\ref{fig:K_analysis} (a) and (b), we can see that when $K$ increases from 5 to 7, the performance does not improve significantly. In practice, we choose $K$ considering both the performance and parameter-efficiency. So, we will choose $K=5$ for Figure~\ref{fig:K_analysis} (a) and (b), and choose $K=7$ for Figure~\ref{fig:K_analysis} (c). 

\subsubsection{Analysis on $\lambda$}

\begin{figure*}[h!]

\vspace{0.1in}

\mbox{\hspace{-0.3in}
     \begin{subfigure}[b]{0.35\textwidth}
         \centering
         \includegraphics[width=\textwidth]{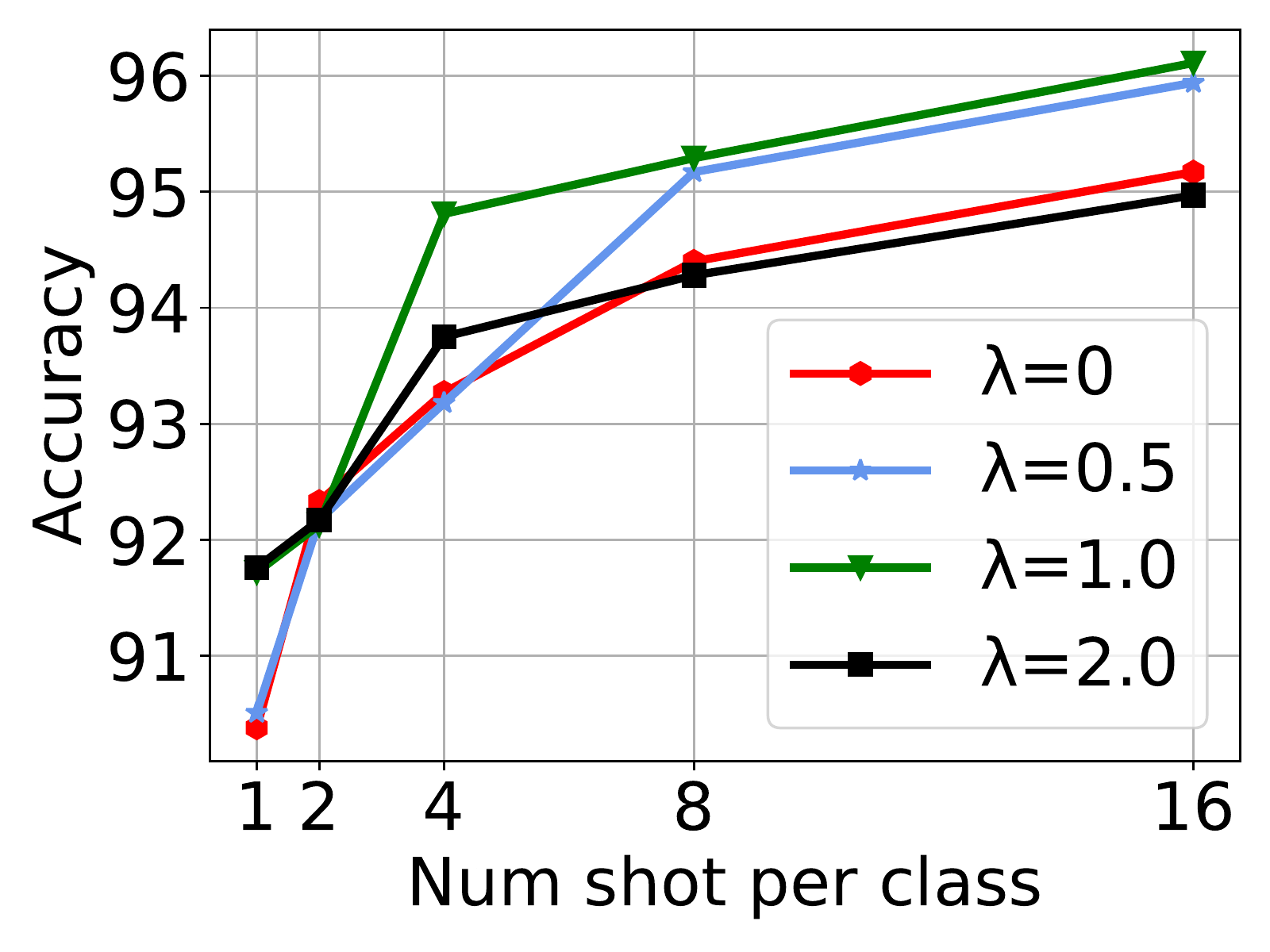}
         \caption{Caltech101 dataset.}
         \label{fig:y equals x}
     \end{subfigure}
     \hspace{-0.1in}
     \begin{subfigure}[b]{0.35\textwidth}
         \centering
         \includegraphics[width=\textwidth]{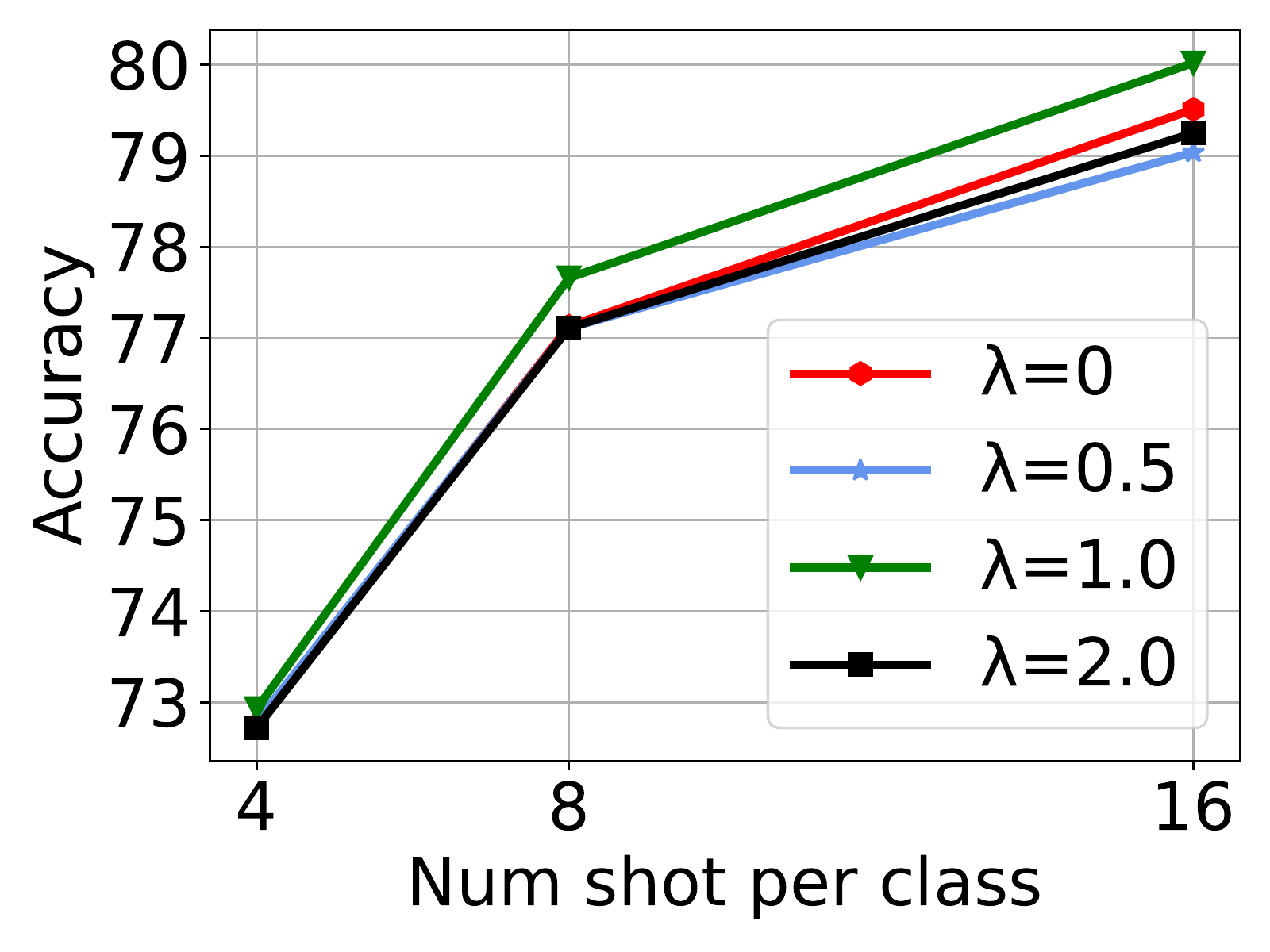}
         \caption{UCF101 dataset.}
         \label{fig:three sin x}
     \end{subfigure}
     \hspace{-0.1in}
     \begin{subfigure}[b]{0.35\textwidth}
         \centering
         \includegraphics[width=\textwidth]{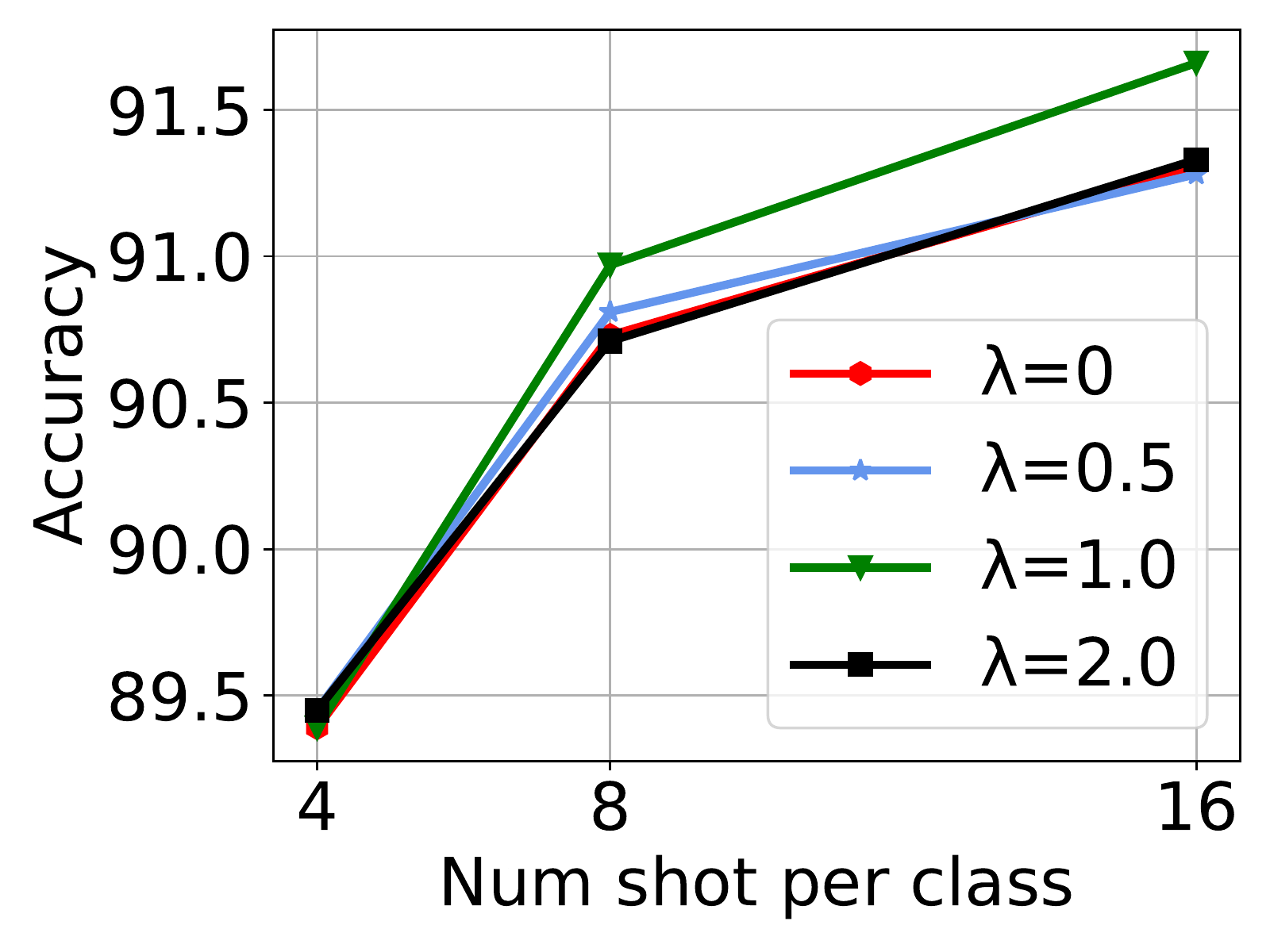}
         \caption{OxfordPets dataset.}
         \label{fig:five over x}
     \end{subfigure}
}
        \caption{Hyper-parameter $\lambda$ analysis on three dataset using CLIP.}
        \label{fig:lambda_analysis}
\end{figure*}

Secondly, we analyze the hyper-parameter $\lambda$, which adjusts the weight of the regularizers. To prove the effectiveness of our regularizers, we set $\lambda=0.0, 0.5, 1.0$, and $2.0$. Setting $\lambda=0$ means we train \texttt{PTP} without regularizers. We conduct experiments using CLIP on datasets Caltech101, UCF101 and OxfordPets, respectively. We report the results in Figure~\ref{fig:lambda_analysis}. When we set the lambda=0.0, this analysis is an ablation study on our defined regularizers, since regularizers are the only parameters we can do an ablation study.

From Figure~\ref{fig:lambda_analysis}, we see that the best performance comes from $\lambda=1.0$. The results prove the significance of our regularizers. Our defined regularizers will push the image prototypes to be meaningful points in the latent space and work as centroids of image clusters. 

\subsubsection{\texttt{PTP} vs. Prompt Learning Baselines}

\begin{figure}[!h]
\centering
\includegraphics[width=0.6\textwidth]{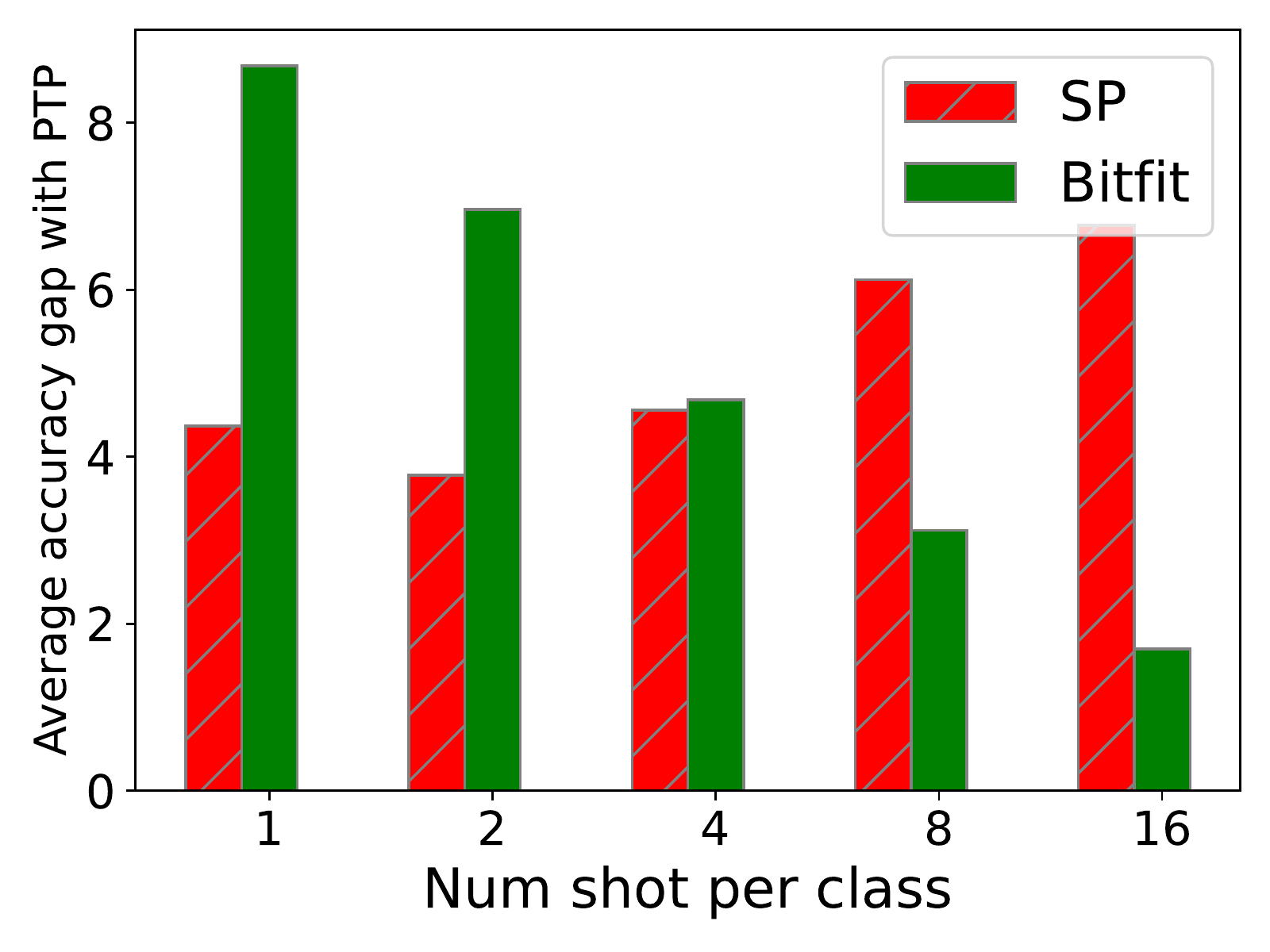}
\vspace{-7pt}
\caption{The average absolute accuracy improvement. Improvement from SP to \texttt{PTP} is shown in red bar, Bitfit to \texttt{PTP} is shown in green bar. Smaller value means smaller gap and better performance. }
\label{fig:bar}
\end{figure}

Generally, we can see that prompt learning baselines SP and CoCoOp are designed only based on one property of PVLM: the aligned images and text (i.e., prompted category names) should have high matching scores. While, our \texttt{PTP} design is also based on the second property of PVLM: similar images are close to each other in the latent space. Leveraging both two properties, our \texttt{PTP} hypothesizes that similar images should use similar prompts. Through prototype-based prompting design, our \texttt{PTP} designs $K$ prompts, which overcomes the drawbacks of task-level method SP and instance-level method CoCoOp. Task-level prompting learns only one prompt for one task, which is suboptimal. Instance-level prompting learns dynamic prompts conditional on instances, which is not lightweight for a few-shot setting. Our \texttt{PTP} outperforms prompt learning baselines on all shots, where the average accuracy gaps between \texttt{PTP} and SP across all the datasets and two PVLMs are shown in Figure~\ref{fig:bar} (Red bar). We see with the shot number increasing, the gap becomes larger. 

\subsubsection{Prompt Learning vs. Parameter-efficient Fine-tuning}

Prompt learning and parameter-efficient fine-tuning methods both have their own applicable scenarios. 

A good prompt learning performance relies on the quality of pre-defined category names. Giving semantic meaningless category names, such as ``707-320'', ``A321'', etc., makes prompt learning methods inferior. On the other hand, the linear probe (LP) does not rely on category names, and Bitfit can update the word embedding and learn semantics through fine-tuning. 

Prompt learning methods highly depend on the PVLM. Prompt learning methods cannot elicit the correct matching if the PVLM itself has limited pre-training visual and text knowledge, since prompting only perturbs the data input. While, with moderate training data, fine-tuning methods can update the image encoding and textual encoding towards optimal. 

Generally, giving a well-trained PVLM with meaningful category names, our \texttt{PTP} is a superior method for few-shot learning, compared with fine-tuning baselines. The average accuracy gaps between \texttt{PTP} and Bitfit fine-tuning are shown in Figure~\ref{fig:bar} (Green bar), where we see with the shot increasing, the gap becomes smaller, but still significant.  

\subsubsection{Limitations}
We summarize two limitations of our model \texttt{PTP}:
i) A good prompt learning performance relies on the quality of pre-defined category names. Hence semantic meaningless category names, such as ``707-320'', ``A321'' and ``BAE-125'', etc., makes prompt learning methods inferior, and ii) Prompt learning methods highly depend on the PVLM. Prompt learning methods cannot elicit the correct matching if the PVLM itself has limited visual and text knowledge during pre-training, since prompting only perturbs the data input. 

\section{Conclusion}
In this work, we propose a prototype-based prompt learning method \texttt{PTP} to overcome the limitations of task-level prompting and instance-level prompting. 
In \texttt{PTP}, the image prototype represents a centroid of a certain image cluster in the latent space and a prompt prototype is defined as a soft prompt in the continuous space. The similarity between a query image and an image prototype determines how much the prediction relies on the corresponding prompt prototype. Hence, in \texttt{PTP}, similar images will utilize similar prompting ways.
We conduct extensive experiments on seven real-world benchmarks for few-shot image recognition task and show that \texttt{PTP} is highly adaptive to various PVLMs with superior performance to other prompt learning methods and parameter-efficient fine-tuning baselines. Moreover, through detailed analysis, we discuss pros and cons of prompt learning v.s. parameter-efficient fine-tuning for few-shot learning. 

\bibliography{refs_scholar}
\bibliographystyle{plainnat}

\end{document}